\documentclass[lettersize,journal]{IEEEtran}
\usepackage{amsmath,amsfonts}
\usepackage{algorithm}
\usepackage{array}
\usepackage[caption=false,font=normalsize,labelfont=sf,textfont=sf]{subfig}
\usepackage{textcomp}
\usepackage{stfloats}
\usepackage{url}
\usepackage{verbatim}
\usepackage{graphicx}
\usepackage{cite}
\usepackage{algpseudocode}

\usepackage{xcolor}
\usepackage{xspace}
\usepackage{bm}
\usepackage{amssymb}
\usepackage{multicol}
\usepackage{booktabs}
\usepackage{pifont}
\usepackage{multirow}

\DeclareMathOperator*{\argmax}{arg\,max}
\DeclareMathOperator*{\argmin}{arg\,min}

\newcommand{\cmark}{\ding{51}}%
\newcommand{\xmark}{\ding{55}}%

\makeatletter
\DeclareRobustCommand\onedot{\futurelet\@let@token\@onedot}
\def\@onedot{\ifx\@let@token.\else.\null\fi\xspace}

\def\eg{\emph{e.g}\onedot} 
\def\ie{\emph{i.e}\onedot} 
 
\def\etc{\emph{etc}\onedot} 
 
\def\etal{\emph{et al}\onedot}
\makeatother

\newcommand{\blue}[1]{{\color{blue}#1}}
\newcommand{\green}[1]{{\color{green}#1}}
\newcommand{\bfy}[1]{\textbf{#1}}

\begin{document}

\title{A Chain-of-Thought Subspace Meta-Learning for Few-shot Image Captioning with Large Vision and Language Models}


\author{
        Hao Huang,
        Shuaihang Yuan,
        Yu Hao,
        Congcong Wen,
        and Yi Fang
\thanks{Hao Huang, Shuaihang Yuan, Yu Hao, Congcong Wen, and Yi Fang are with NYU Embodied AI and Robotics Lab, New York University, USA, and New York University Abu Dhabi, UAE. Email:\{hh1811,sy2366,yh3252,cw3437,yfang\}@nyu.edu.}

% \thanks{$\dagger$ Corresponding author: Yi Fang (yfang@nyu.edu).}
}

\markboth{Journal of \LaTeX\ Class Files,~Vol.~14, No.~8, August~2021}%
{Shell \MakeLowercase{\textit{et al.}}: A Sample Article Using IEEEtran.cls for IEEE Journals}


\maketitle

\begin{abstract}
A large-scale vision and language model that has been pretrained on massive data encodes visual and linguistic prior, which makes it easier to generate images and language that are more natural and realistic. Despite this, there is still a significant domain gap between the modalities of vision and language, especially when training data is scarce in few-shot settings, where only very limited data are available for training. In order to mitigate this issue, a multi-modal meta-learning framework has been proposed to bridge the gap between two frozen pretrained large vision and language models by introducing a tunable prompt connecting these two large models. For few-shot image captioning, the existing multi-model meta-learning framework utilizes a one-step prompting scheme to accumulate the visual features of input images to guide the language model, which struggles to generate accurate image descriptions with only a few training samples. Instead, we propose a chain-of-thought (CoT) meta-learning scheme as a multi-step image captioning procedure to better imitate how humans describe images. In addition, we further propose to learn different meta-parameters of the model corresponding to each CoT step in distinct subspaces to avoid interference. We evaluated our method on three commonly used image captioning datasets, \ie, MSCOCO, Flickr8k, and Flickr30k, under few-shot settings. The results of our experiments indicate that our chain-of-thought subspace meta-learning strategy is superior to the baselines in terms of performance across different datasets measured by different metrics.
\end{abstract}

\begin{IEEEkeywords}
Few-shot learning, Image captioning, multi-modality foundation models, Vision and language models, Chain-of-Thought, Subspace learning.
\end{IEEEkeywords}

%%%%%%%%%%%%%%%%%%%%%%%%%%%%%%%%%%%%%%%%%%%%%%%%%%%%%%%%%%%%%%%%%%%%%%%%%%%%%%%%%%%%%%%%%
\section{Introduction}
\label{sec:intro}
\IEEEPARstart{M}{eta-learning}, often referred to as ``learning to learn'', is a burgeoning field that seeks to enhance the efficiency of the learning process by acquiring meta-knowledge from a variety of different tasks~\cite{hospedales2021meta,li2021concise}. The primary objective of meta-learning is to design algorithms or models that can adapt to unseen new tasks effectively, especially even when training data are limited. This objective is usually achieved by training a meta-learner on different tasks, each of which comprises a training set and a validation set. The meta-learner optimizes its parameters based on performance on the validation set, enabling it to adapt quickly to unseen new tasks to make accurate predictions. This approach is particularly beneficial in few-shot learning scenarios, where there is a scarcity of training data for each task~\cite{santoro2016meta,wang2020generalizing}.

\begin{figure}[!htb]
    \centering
    \includegraphics[width=0.99\linewidth]{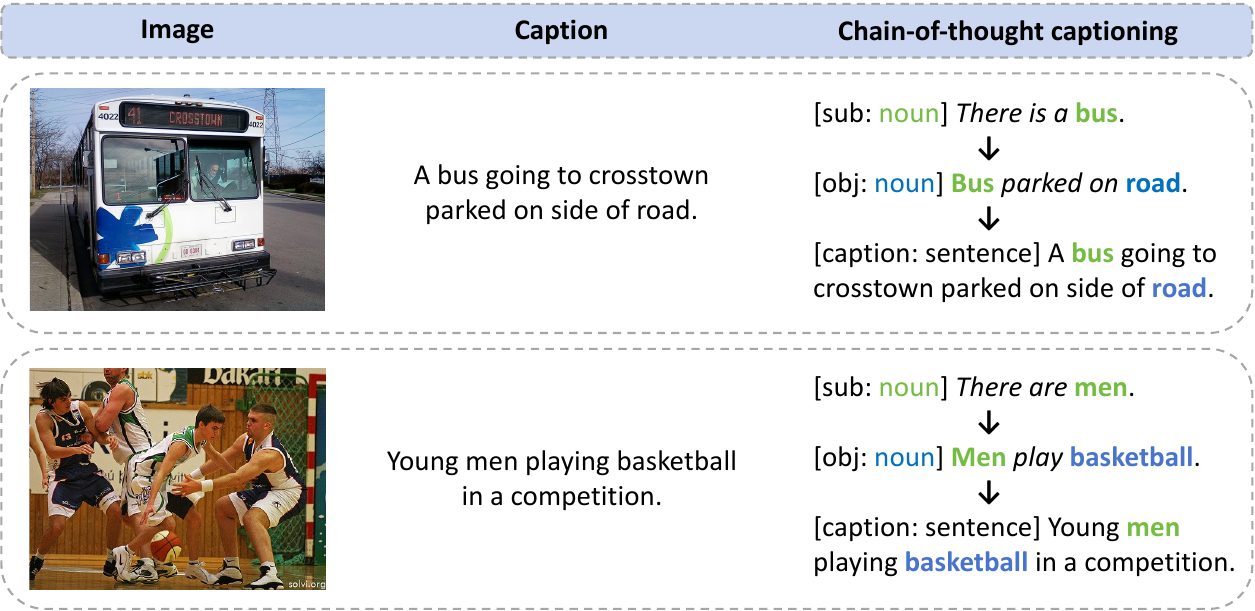}
    \caption{By prompting an LLM sequentially the subject (sub: \green{bus}/\green{men}) and object (obj: \blue{road}/\blue{basketball}) contained in an image, the LLM could reason about their interactions/relations (verb: parked/playing) to generate an accurate caption for the image.}
    \label{fig:teaser}
\end{figure}

% \red{Introduce large language model and large vision model.} \\
Due to the limitations of hardware and algorithms, previous deep meta-learning approaches focus mainly on optimizing small- to medium-scale neural networks~\cite{ravi2016optimization,finn2017model,grant2018recasting}, 
Recently, large language models (LLMs), designed to comprehend and generate natural human language, have gained substantial attention due to their ability to produce coherent and contextually relevant text~\cite{radford2019language,brown2020language}. Trained on massive text corpus, these models learn patterns and relationships within languages, and have been applied in various domains of natural language processing, including but not limited to, machine translation~\cite{zhu2023multilingual}, text generation~\cite{xiao2023patterngpt}, and question answering~\cite{robinson2022leveraging}, \etc.  Similarly, large vision models (LVMs), also commonly known as \textit{vision Transformers}, have demonstrated impressive performance in numerous vision tasks~\cite{beal2022billion}. These models, based on the Transformer architecture~\cite{vaswani2017attention} initially introduced for natural language processing, leverage self-attention mechanisms to capture global dependencies within image patches, enabling them to understand the relationships between different regions and objects within an image~\cite{dosovitskiy2020image}. Pretraining these models on massive image datasets allows them to learn rich visual representations such that they can be directly used or fine-tuned for specific downstream tasks, achieving state-of-the-art performance on various computer vision benchmarks~\cite{wang2022bevt,dehghani2023scaling,singh2023effectiveness}. Meanwhile, the downside of training LLMs and LVMs is that this process requires substantial computational resources, leading to very high time and hardware costs.

Despite the computational challenges for LLM and LVM training, the pretrained LLM and LVM have significantly advanced the field of natural language processing and computer vision.  For example, Brown \etal~\cite{brown2020language} found that by providing a few-shot demonstrations specified purely by text to GPT-3, it can perform most language tasks without any gradient updates or fine-tuning required by the conventional neural network models. However, how to extend the few-shot learning ability to multi-modality models containing both an LLM and an LVM is still challenging, due to the discrepancy between the embedding spaces of the LLM and LVM. Previous work~\cite{tsimpoukelli2021multimodal} only updated the LVM's parameters to learn visual features while freezing the LLM's parameters, and~\cite{najdenkoska2022meta} instead introduced a light-weighted learnable meta-mapper to tune visual features generated by a frozen pretrained LVM to bridge the gap between the two embedding spaces of the frozen LLM and LVM for few-shot visual question answering. 

% \red{Introduce prompt and chain of thought.} \\
The aforementioned works~\cite{tsimpoukelli2021multimodal,najdenkoska2022meta} regard learned or tuned visual features as ``\textit{prompt}''~\cite{wei2022emergent} to guide an LLM in generating output for specific tasks, \eg, image captioning or visual question answering, inspired by the fact that prompt can be used as conditions to provide contexts for the LLM to generate responses~\cite{lester2021power}.  However, for complex multi-modality tasks, we speculate that \textit{a single-stage prompting may not be enough to guide an LLM to generate accurate responses}. Taking image captioning as an example, only prompting a single global visual feature of an input image to an LLM does not fully utilize language prior knowledge stored in the LLM: an image usually contains several interactional objects, but the LLM may not generate accurate descriptions to describe these objects and reason about their interactions without explicitly identifying and attending to them.

Chain-of-Thought (CoT) prompting~\cite{wei2022chain} is a strategy aimed at eliciting reasoning in LLMs. It involves inserting \textit{multi-stage reasoning paths before generating the final response}. This approach allows an LLM to engage in a sequence of logical steps to arrive at the desired answer, promoting more sophisticated reasoning abilities~\cite{saparov2022language} to improve its performance on tasks that require logical thinking~\cite{kojima2022large}.  We refer the interested reader to~\cite{qiao2022reasoning} for a more comprehensive review of prompt and chain-of-thought. In this work, we explore CoT in bridging an LLM and an LVM for multi-modal few-shot image captioning in a meta-learning framework, as the process of humans describing an image is also a logical thinking process entails identifying objects within the image and discerning their relationships, thereby constructing a coherent narrative based on visual cues and spatial arrangement. Specifically, we sequentially prompt an LLM with a visual feature of a \textit{subject} in an image, a visual feature of an \textit{object} in the image, and finally a global visual feature of the image to generate the final caption in few-shot scenarios, as conceptually illustrated in Figure~\ref{fig:teaser}.  We assume that by conditioning on the subject and object in an image, an LLM could reason about their interactions based on the language prior stored in it and then produce an accurate description.  The CoT learning process is embedded in a meta-learning framework, \ie, MAML~\cite{finn2017model}. Furthermore, since different steps in CoT extract different information from an image as prompts, we further propose a subspace parameter structure in which \textit{each subspace embodies a distinct form of meta-knowledge of the corresponding step of CoT}, with the bases of these subspaces as meta-parameters. For every step of CoT, the meta-learner constructs a prompt-specific model derived from each subspace. Subsequently, the meta-learner refines the bases of these subspaces by reducing the weighted validation loss associated with the task models. We evaluated our method on three widely used image captioning datasets, \ie, MSCOCO~\cite{lin2014microsoft}, Flickr8k~\cite{hodosh2013framing}, and Flickr30k~\cite{young2014image}, in few-shot scenarios, and the experimental results demonstrate that our method outperforms the compared approaches.

% \red{Introduce our method: 1) Chain of thought; 2) subspace.} \\

%%%%%%%%%%%%%%%%%%%%%%%%%%%%%%%%%%%%%%%%%%%%%%%%%%%%%%%%%%%%%%%%%%%%%%%%%%%%%%%%%%%%%%%%%
\section{Related Work}
\label{sec:related}
\noindent \textbf{Large-scale language and vision models.} 
The advent of attention mechanisms~\cite{bahdanau2015neural} and Transformers~\cite{vaswani2017attention} marked the emergence of large-scale language models, effectively addressing the challenge of long-range dependencies in sequences, including but not limited to, BERT~\cite{devlin2019bert}, GPT-2~\cite{radford2019language}, GPT-3~\cite{brown2020language}, ChatGPT~\cite{openai2023chatgpt} and GPT-4~\cite{openai2023gpt4}. BERT is designed to understand bidirectional contexts using the Transformer's encoder component to read the entire input text at once and learn the relationships between different words in the text; GPT-based models are primarily trained to predict the next word in a sentence given all the previous words by using the Transformer's decoder component to process text unidirectionally~\cite{ghojogh2020attention}. Meanwhile, large vision models also emerge in computer vision and multi-modal fields. For example, Vision Transformer (ViT)~\cite{dosovitskiy2020image} has broadened the use of the Transformer architecture to establish an analogy between word tokens and image patches. Subsequent efforts have sought to improve and expand the ViT model, as evidenced by Swin Transformer~\cite{liu2021swin}, MAE~\cite{he2022masked}, IPT~\cite{chen2021pre}, and BeiT~\cite{bao2021beit}, \etc. These studies have effectively adapted the ViT model to a variety of vision-related tasks, yielding exceptional results. Simultaneously, studies in the multi-modal domain have embraced Transformer for cross-modal data interaction, including CLIP~\cite{radford2021learning} for text-image alignment, Coca~\cite{yu2022coca} and ClipCap~\cite{mokady2021clipcap} for image captioning, DALL-E~\cite{ramesh2021zero} for text-to-image generation, and PALI~\cite{wang2023image} for visual question answering, \etc. In addition, BLIP~\cite{li2022blip} utilizes the noisy web data by bootstrapping the captions to train a large language-vision model, and BLIP-2~\cite{li2023blip} bridges the gap between multi-modalities with a lightweight Querying Transformer. We refer the interested reader to~\cite{zhao2023survey} and~\cite{wang2023review} for more literature.

\noindent \textbf{Prompt and chain-of-thought.}
Prompting~\cite{brown2020language} is to guide large pretrained LLMs towards generating the desired response by treating them as knowledge bases from which information useful to downstream tasks is elicited~\cite{petroni2019language}. This methodology involves the use of a few examples as prompts, appended to a fixed task induction, to stimulate the LLM to generate accurate output. Such prompts are often designed manually by the user. Prompt tuning was initially introduced in~\cite{lester2021power}, which instead of manually designing a prompt, learns ``soft prompts'' to condition frozen LLMs to perform specific downstream tasks, while preserving the embedding space structure of the pretrained LLMs~\cite{gao2021making}. Recent studies, such as L2P~\cite{wang2022learning} and AutoPrompt~\cite{shin2020autoprompt}, selected the most effective prompts from a collection of learned potential prompts. The concept of prompt tuning has also gained significant interest in computer vision. Adaptformer~\cite{chen2022adaptformer} employed prompt tuning to efficiently adapt pretrained ViTs, while CoOp~\cite{zhou2022learning} and CoOpOp~\cite{zhou2022conditional} applied prompt tuning to adapt a pretrained large visual-language model by transforming the context word into a set of learnable vectors. Recently, Chain-of-Thought (CoT) prompting has been widely used to extract the multi-step intermediate reasoning capabilities of LLMs~\cite{wei2022chain}. Specifically, CoT stimulates an LLM to generate intermediate reasoning chains to solve complex problems. For instance,~\cite{huang2023inner} and~\cite{lu2022learn} proposed using the reasoning process by language models for robot interaction and question answering. EmbodiedGPT~\cite{mu2024embodiedgpt} proposed an end-to-end multi-modal foundation model for embodied AI by using CoT for effective embodied planning. Despite massive empirical success, the most recent work~\cite{feng2024towards} revealed the underlying mechanisms behind CoT using circuit complexity theory.

\noindent \textbf{Meta-learning and few-shot learning.} 
The objective of meta-learning is to learn to acquire inductive biases and perform rapid adaptation to novel unseen tasks~\cite{grant2018recasting}. Meta-learning algorithms are typically classified into three categories: (i) Metric-based approaches which focus on learning a common embedding space and generating prototypes as meta-knowledge~\cite{vinyals2016matching,snell2017prototypical,sung2018learning}; (ii) Memory-based approaches which utilize an external memory as meta-knowledge to quickly adapt to new tasks~\cite{santoro2016meta,munkhdalai2017meta,munkhdalai2018rapid,mishra2018simple}. (iii) Optimization-based approaches that aim to learn a good model initialization across tasks as meta-knowledge to efficiently adapt to new samples~\cite{ravi2016optimization,finn2017model,grant2018recasting}. Meta-learning has been proved to be effective in solving few-shot learning tasks by allowing models to generalize from only a small number of training samples~\cite{elsken2020meta,huang20213d,lang2023base}. This approach trains models on a variety of learning tasks so that they can quickly adapt to new tasks using limited data, mimicking human learning efficiency. The field of multimodal few-shot learning, spanning vision and language modalities, has recently gained traction with the advent of the first multimodal few-shot learner~\cite{tsimpoukelli2021multimodal}. This development was followed by other innovative prompting and in-context learning strategies~\cite{jin2022good,song2022clip}. Flamingo~\cite{alayrac2022flamingo} utilized a super large visual-language model comprising 70 billion parameters for in-context learning. MiniGPT-4~\cite{zhu2023minigpt} aligns a frozen visual encoder with a frozen LLM through a single projection layer. However, our objective diverges from the above multi-modality visual-language models, as we aim to rapidly adapt to new unseen tasks by acquiring meta-knowledge across different tasks as in~\cite{najdenkoska2022meta}, by incorporating a light-weighted model with much fewer trainable parameters. 

\noindent \textbf{Image captioning.} 
The image captioning task involves generating descriptive text for an image to interpret and articulate the content and context of visual data contained in the image±\cite{vinyals2015show,xu2015show}. This task can usually be tackled in two strategies: retrieval-based and generation-based. Retrieval-based approaches~\cite{zhao2020cross,al2022image} retrieve complete sentences that best describe the given image from a corpus, while generation-based approaches~\cite{vinyals2015show,xu2015show} construct sentences by generating words in sequence, which typically contain a visual encoder to extract image features and a sequence-to-sequence model such as LSTM~\cite{hochreiter1997long} and/or Transformer~\cite{vaswani2017attention} for text generation. To extract accurate image features, previous work~\cite{you2016image,yang2022human,wang2019hierarchical} proposed using Region-of-Interest (RoI) features generated from off-the-shelf objector detectors~\cite{girshick2015fast}. For text generation, previous approaches~\cite{anderson2016spice,yao2019hierarchy,pan2020x} typically utilize LSTM, while subsequent work~\cite{li2020oscar,wang2022end,zeng2022s2} leverages attention-based models such as Transformers to predict captions. Recent studies have integrated the retrieved image-caption pairs into generation-based models~\cite{sarto2022retrieval,ramos2023smallcap}, which aligns with the concept of in-context captioning that emerges in the era of LLMs. To utilize the generalization ability of pretrained LLMs, the weights of LLMs are partially or completely frozen to prevent catastrophic forgetting~\cite{zhai2024investigating}. ClipCap~\cite{mokady2021clipcap} and ITuning~\cite{luo2022tuning} are two lightweight tunable image captioning models that use CLIP~\cite{song2022clip} as a pre-trained vision encoder and GPT-2~\cite{radford2019language} as a language decoder. Though the components of CLIP and GPT-2 are frozen, ClipCap employs prefix-tuning to map a fixed-length CLIP embedding of an input image into the GPT-2 language space, and I-Tuning extracts visual memory embeddings from CLIP and uses those to adjust the output hidden states of GPT-2. SMALLCAP~\cite{ramos2023smallcap} uses retrieval augmentation to maintain performance while substantially reducing the number of trainable parameters. Instead of randomly selecting in-context image-caption pairs, Yang \etal~\cite{yang2024exploring} devised a strategy for image captioning by selecting images and assigning captions to establish in-context image-caption pairs.

%%%%%%%%%%%%%%%%%%%%%%%%%%%%%%%%%%%%%%%%%%%%%%%%%%%%%%%%%%%%%%%%%%%%%%%%%%%%%%%%%%%%%%%%%
%
\begin{figure*}[!htb]
    \centering
    \includegraphics[width=0.99\linewidth]{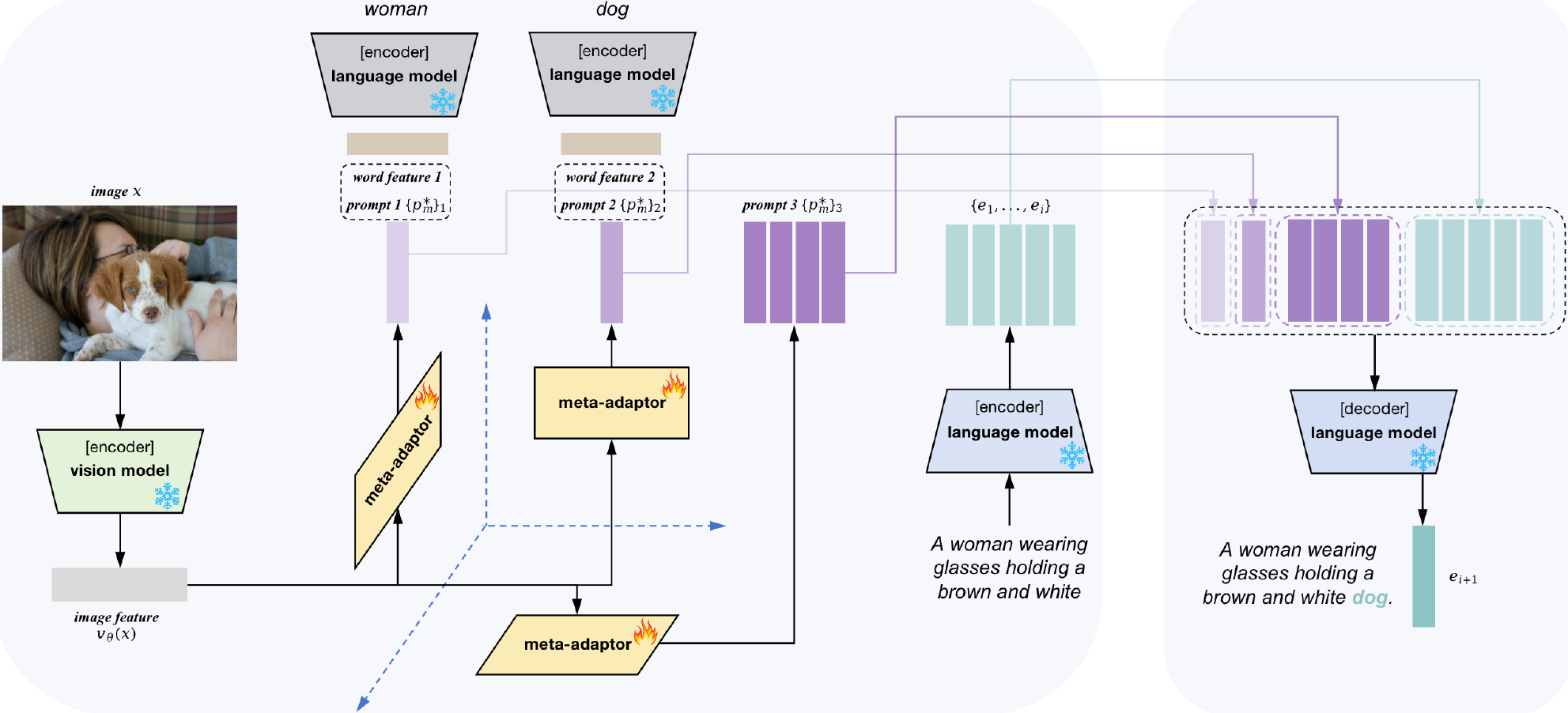}
    \caption{Model architecture overview. Given an image, a vision encoder first extracts an image feature which is then transformed into multiple visual prompts by meta-adaptors in different subspaces through meta-learning; these visual prompts
are sequentially fed into an LLM along with word tokens to generate a caption.}
    \label{fig:arch}
\end{figure*}

\section{Method}
\label{sec:method}
This section describes our multi-modal chain-of-thought subspace meta-learning for few-shot image captioning. We first outline the problem of few-shot image captioning in Section~\ref{subsec:problem}; then we illustrate our meta-learning vision-language model architecture for few-shot image captioning in Section~\ref{subsec:arch}, followed by a description of chain-of-thought caption generation procedure in Section~\ref{subsec:cot}; and finally, we detail meta-learing in subspaces in Section~\ref{subsec:subspace}. Our goal is to develop a vision-language model for image captioning such that it can swiftly adjust to new captioning tasks with scarce image-caption paired data within the context of meta-learning. 

% ------------------------------------------------------------------------------------
\subsection{Problem Formulation}
\label{subsec:problem}
Contrasting with traditional supervised learning, meta-learning encompasses a collection of meta-datasets, which are partitioned into disjoint \textit{meta-train} and \textit{meta-test} sets. In a \textit{meta-train} stage, the model is trained on the meta-train set, and in a separate \textit{meta-test} stage, the model is evaluated on the meta-test set. Specifically, the meta-train set is composed of meta-datasets $\{\mathcal{D}_n^{tr}\}$, each containing a pair of distinct inner-train set (or \textit{support set}) $\{\mathcal{S}_n\}$, and inner-test set (or \textit{query set}) $\{\mathcal{Q}_n\}$. Thus, the meta-train set is defined as $\{(\mathcal{S}_1, \mathcal{Q}_1), \cdots, (\mathcal{S}_n, \mathcal{Q}_n), \cdots\}$. We denote each pair $(\mathcal{S}_n, \mathcal{Q}_n)$ as a task $\mathcal{T}_n$~\cite{finn2017model}. Within a few-shot $N$-way $K$-shot setting, $\mathcal{S}_n$ for an individual task $\mathcal{T}_n$ includes $K$ (labeled) samples from each of the $N$ object categories or classes. This suggests that $\mathcal{S}_n = \{(x^n_1, y^n_1), \cdots, (x^n_{KN}, y^n_{KN})\}$, where $x^n_j$ represents the image and $y^n_j$ denotes the corresponding caption. Similarly, the query set is characterized by $\mathcal{Q}_n = \{(x^n_1, y^n_1), \cdots, (x^n_{LN}, y^n_{LN})\}$, where $L$ is the number of samples per category or class of objects, and could be different from $K$. Given a task distribution $p(\mathcal{T})$, we first sample a batch of tasks $\{\mathcal{T}_n\} \sim p(\mathcal{T})$, then $N$ object categories are randomly drawn for each task $\mathcal{T}_n$, and finally $K$ and $L$ samples, \ie, image-caption pairs, are selected at random for each of the $N$ categories to construct $\mathcal{S}_n$ and $\mathcal{Q}_n$. The meta-test set $\{\mathcal{D}_m^{te}\} = \{(\mathcal{S}_1, \mathcal{Q}_1), \cdots, (\mathcal{S}_m, \mathcal{Q}_m), \cdots\}$ is constructed in a similar way. Note that in a few-shot setting, the value $N$ is relatively small, usually set between 1 and 5.

Using the above notation, the meta-train stage can be formally written as~\cite{hospedales2021meta}:
\begin{equation}
\omega^\ast = \argmin_\omega\log p(\omega \mid \{\mathcal{D}_n^{tr}\})\enspace,
\label{eq:meta_train}
\end{equation}
where $\omega$ is defined as \textit{across-task} knowledge or \textit{meta-knowledge} learned from the meta-train set. In the meta-test stage, the learned meta-knowledge is used to train a predictive model for captioning on each previously unseen task. For the $m$-th meta-test task, we obtain the optimal model parameters conditioned on the meta-knowledge as:
\begin{equation}
    \theta^\ast_m = \argmax_\theta \log p(\theta \mid \omega^\ast, \mathcal{S}_m)\enspace,
    \label{eq:meta_test}
\end{equation}
and the performance of $\theta^\ast_m$ is evaluated on $\mathcal{Q}_m$ where $(\mathcal{S}_m, 
\mathcal{Q}_m) \in \{\mathcal{D}_m^{te}\}$ as defined above. Here, $\omega$ could be an initial neural network parameters~\cite{finn2017model} as in our case, a parameterized loss function~\cite{li2019feature} or a hyper-parameter in the regularization term~\cite{franceschi2018bilevel}.

% ------------------------------------------------------------------------------------
\subsection{Model Architecture}
\label{subsec:arch}
As shown in Figure~\ref{fig:arch}, our model consists of three components: a frozen LVM encoder, multiple learnable meta-adaptors, and a frozen LLM with an encoder and a decoder. Given an input image, the LVM encoder first extracts the image feature which is then transformed into multiple visual prompts through the meta-adaptors; these visual prompts are sequentially fed into an LLM to generate a caption. The meta-adpators parameters are learned through meta-learning in different subspaces.

\noindent \textbf{Vision encoder.}
We denote a pretrained LVM encoder~\cite{dosovitskiy2020image} as $v_\phi$, with the parameters $\phi$ fixed during training. Given an image $x$, the vision encoder extracts the feature $v_\phi(x)$ of the given image.

\noindent \textbf{Meta-adaptor.} 
For the $k$-th step in CoT (described in the next subsection), to convert the image feature $v_\phi(x)$ into the latent space of the LLM as a visual prompt to guide the LLM to generate captions, we utilize a set of $c$ learnable parameters $\{p_1, \cdots, p_c\}_k$\footnote{For conciseness, we use $\{p_{1\cdots c}\}$ to denote $\{p_1, \cdots, p_c\}$.}, prepended to the image feature, resulting in a sequence $[\{p_{1\cdots c}\}_k, v_\phi(x)]$ in the $k$-th step. Then, we apply multi-head attention block~\cite{lee2019set,mokady2021clipcap,najdenkoska2022meta} to encode the entire sequence concurrently, which acts as a \textit{meta-adaptor} $f_\theta$ with trainable meta-parameters $\theta$, and thus the attention block is defined as:
\begin{equation}
 f_\theta(Q, K, V) = \sigma(QK^\top)V\enspace,   
\label{eq:qkv}
\end{equation}
where $\sigma$ represents a softmax function. In Eq.~\ref{eq:qkv}, we set $Q = K = V = [\{p_{1\cdots c}\}_k, v_\phi(x)]$, and the output is a vector of learned parameters $\{p^\ast_{1\cdots c}\}_k$ expressed as:
\begin{equation}
 \{p^\ast_{1\cdots c}\}_k = f_\phi([\{p_{1\cdots c}\}_k, v_\phi(x)])\enspace.  
\label{eq:mapper}
\end{equation}
The self-attention block facilitates the extraction of meaningful information from the image feature $v_\phi(x)$, and accumulates it into multiple visual prompts $\{p^\ast_{1\cdots c}\}_k$. The meta-parameters of the meta-adaptor are trained and shared across all tasks in the meta-train set $\{\mathcal{D}_n^{tr}\}$. As the number of parameters of the meta-adaptor is several orders of magnitude smaller than that of the LVM, the training process of the meta-adaptor is both rapid and efficient, facilitating faster convergence in the scarcity of training data per task.

\noindent \textbf{Language model.} 
The language model~\cite{radford2019language} is denoted as $l_\psi$ which is parametrized by $\psi$ and models a probability distribution over a text sequence $y$. Precisely, to predict the $i$-th token in a caption, the language model receives a chain of visual prompts $\bigl\{\{p^\ast_{1\cdots c}\}_1, \cdots, \{p^\ast_{1\cdots c}\}_k\bigr\}$ concatenated with the token embeddings $\{e_1, \cdots, e_{i-1}\}$, and produces the following token in an auto-regressive fashion:
\begin{equation}
e_i = l_\psi([\{p^\ast_{1\cdots c}\}_1, \cdots, \{p^\ast_{1\cdots c}\}_k, e_1, \cdots, e_{i-1}])\enspace.
\label{eq:gpt}
\end{equation}
As we can only access to limited samples per task under a few-shot scenario, we initialize the $\psi$ parameters with pretrained parameters from an LLM~\cite{radford2019language} and freeze them during training. This initialization acts as a language prior, embedding a vast knowledge base into the LLM from the outset, and these parameters are kept frozen throughout the training process to leverage the LLM's generalization capabilities without overfitting to the scarce task-specific data.

\label{subsec:train}

% ------------------------------------------------------------------------------------
\subsection{Chain-of-Thought: Prompt Chaining}
\label{subsec:cot}
We propose to decompose image captioning procedure into a sequential cognitive reasoning mechanism. To emulate cognitive reasoning, we establish a series of causal prompts as a chain, with each prompt receiving and transmitting information in sequence. To achieve CoT image captioning that emulates human-like captioning procedure, we utilize the concept of \textit{Subject-verb-Object} (SVO) to serve as a reasoning mechanism for caption generation. As depicted in Figure~\ref{fig:teaser}, an SVO is formally defined as:
\begin{equation}
    \text{SVO} = \{\texttt{sub}, \textit{verb}, \texttt{obj}\}\enspace,
\end{equation}
where the \textit{verb} represents a \textit{predicate} that encapsulates a main action/activity within an image; \texttt{sub} denotes a \textit{subject} role associated with verb and \texttt{obj} represents an \textit{object} role corresponding to the subject. In linguistic typology, the SVO structure indicates that the subject comes first, the verb second, and the object third in a sentence\footnote{We regard both ``A child plays football.'' and ``A cat is on a table.'' fall into this format. Statistically, SVO and its SOV variant account for more than 87\% of the languages of the world~\cite{crystal1997cambridge}.}. SVO offers a precise mechanism to control the granularity of information embedded in the image sentence. The construction of SVOs can be achieved either automatically or through manual annotation. With regard to that \texttt{sub} and \texttt{obj} are generally nouns that refer to objects in an image, they can be identified with high precision using an existing pretrained image classification model or image tagging model. As for \textit{verb}, since there is no pretrained action or activity recognition model available for accurate image action or activity recognition, we simply generate it in captions by leveraging the language prior stored in a pretrained LLM.

Given an image $x$, the objective is to describe $x$ through a textual sequence $y$, essentially estimating the probability $p(y|x)$. Drawing inspiration from human cognitive reasoning in image description discussed above, we decompose this task into three steps of CoT: generate \texttt{sub} first, then predict \texttt{obj}, and finally populating \texttt{sub} and \texttt{obj} to generating a caption containing a \textit{verb} that relates the \texttt{sub} and the \texttt{obj}:
\begin{equation}
    p(y|x) = p(\texttt{sub}|x)p(\texttt{obj}|\texttt{sub},x)p(y|\texttt{obj},\texttt{sub},x)\enspace.
    \label{eq:cot}
\end{equation}
The three terms on the right-hand side in Eq.~\ref{eq:cot} form a chain with $k=3$ reasoning steps. Through this approach, every step assimilates the results of the previous reasoning and engages in its own cognitive process for the current step. As a result, our sequential prompts maintain an intrinsic reasoning capability for captioning. 
% \red{Link with notations in Section~\ref{subsec:arch}.}

% ------------------------------------------------------------------------------------
\subsection{Subspace MAML}
\label{subsec:subspace}
Model-Agnostic Meta-Learning (MAML)~\cite{finn2017model} is designed to optimize the parameters of a model so that it can rapidly adapt to novel or new tasks with minimal additional training. In the context of MAML, meta-parameters refer to the initial set of parameters from which the model starts before it adapts to a new task. The goal is to find a set of initial parameters that allow the model to learn a new task quickly and effectively with only a small number of gradient steps and limited data. Thus, these meta-parameters serve as a versatile starting point that can be fine-tuned for various specific tasks with minimal additional training. Note that all the meta-parameters in MAML are learned in a shared parameter space. However, since each step in the chain captures distinct levels of information in an image, \ie, subject level $\leftrightarrow p(\texttt{sub}|x)$, subject $+$ object level $\leftrightarrow p(\texttt{obj}|\texttt{sub},x)$, and image content level $\leftrightarrow p(y|\texttt{obj},\texttt{sub},x)$, it is unreasonable to learn the meta-parameters in a shared parameter space. Therefore, inspired by~\cite{jiang2022subspace}, we propose to learn these meta-parameters of each CoT step in different subspaces.

Taking a single layer in a meta-adaptor as an example, which can be easily extended to all layers, we assume that the meta-parameter of this layer at the $k$-th step ($k \in \{1, \cdots, K\}$) in CoT is represented as $\bm{w}_k \in \mathbb{R}^{d\times p}$, in which each column $\bm{w}_k[:, i] \in \mathbb{R}^d$ resides within a subspace $\mathbb{S}_k$. Furthermore, we posit that each subspace $\mathbb{S}_k$ has the same dimensionality $D$. Let $\bm{S}_k \in \mathbb{R}^{d\times D}$ be the basis for $\mathbb{S}_k$ where each column $\bm{S}_k[:, j] \in \mathbb{R}^d$ represents a base vector. Thus, $\bm{w}_k$ can be calculated as a set of linear combinations of columns of $\bm{S}_k$:
\begin{equation}
\bm{w}_k = \bm{S}_k \bm{c}_k\enspace,
\label{eq:comb}
\end{equation}
where the coefficients $\bm{c}_k \in \mathbb{R}^{D \times p}$. The set $\{\bm{c}_1, \cdots, \bm{c}_K\}$ and $\{\bm{S}_1, \cdots, \bm{S}_K\}$ constitute the meta-parameters to be learned. 

Formally, at the $k$-th step in CoT, the base learner searches for the model parameter $\bm{w}_k$ in the fixed subspace $\bm{S}_k$. Within each subspace $\mathbb{S}_k$, the optimal linear combinations $\bm{c}_k^\ast$ of the subspace's basis is defined as: 
\begin{equation}
    \bm{c}_k^\ast = \argmin_{\bm{c}_k \in \mathbb{R}^{D \times p}}\mathcal{L}(\{\mathcal{D}_n^{tr}\}; \bm{S}_k\bm{c}_k)\enspace.
\end{equation}
The resultant $\bm{S}_k\bm{c}_k^\ast$ represents the CoT  meta-parameters associated with the subspace at the $k$-th step in COT. We approximate the solution of $\bm{S}_k\bm{c}_k^\ast$ by executing gradient descent iterations as in MAML and its variant~\cite{finn2017model,jiang2022subspace}, starting from an initial value:
\begin{equation}
    \bm{c}_k^\prime = \bm{c}_k - \alpha \nabla_{\bm{c}_k}\mathcal{L}(S_n; \bm{S}_k\bm{c}_k)\enspace,
\end{equation}
where $\alpha$ denotes a positive learning rate. The learning process is described in Algorithm~\ref{alg:meta}.

\begin{algorithm}
\caption{Chain-of-Thought subspace meta-learning}
\begin{algorithmic}[1]
\Require Task distribution $p(\mathcal{T})$; positive learning rate $\alpha$ and $\beta$ for inner and outer loop; the number $K$ of steps in CoT; initialization $\{\bm{S}_k\}$ and $\{\bm{c}_k\}$
\While {\textit{not} done}
    \State Sample a batch of tasks $\{\mathcal{T}_n\}$ from $\mathcal{T}$
    \For {each task $\mathcal{T}_n$ in $\{\mathcal{T}_n\}$}
        \State Sample $(S_n, Q_n)$ from $\mathcal{T}_n$
        \For {each step $k \{1, \cdots, K\}$ in CoT}
            \State Compute $\nabla_{\bm{c}_k}\mathcal{L}(S_n; \bm{S}_k\bm{c}_k)$
            \State $\bm{c}_{k,n}^\prime \leftarrow \bm{c}_k - \alpha \nabla_{\bm{c}_k}\mathcal{L}(S_n; \bm{S}_k\bm{c}_k)$
        \EndFor
        \For {each step $k \in \{1, \cdots, K\}$ in CoT}
            \State Compute $\nabla_{\bm{c}_k}\mathcal{L}(Q_n; \bm{S}_k\bm{c}_{k,n}^\prime)$
            \State Compute $\nabla_{\bm{S}_k}\mathcal{L}(Q_n; \bm{S}_k\bm{c}_{k,n}^\prime)$
        \EndFor
    \EndFor
    \For {each step $k \in \{1, \cdots, K\}$ in CoT}
        \State Update $\bm{c}_k \leftarrow \bm{c}_k - \beta\sum_n \nabla_{\bm{c}_k}\mathcal{L}(Q_n; \bm{S}_k\bm{c}_{k,n}^\prime)$
        \State Update $\bm{S}_k \leftarrow \bm{S}_k - \beta\sum_n \nabla_{\bm{S}_k}\mathcal{L}(Q_n; \bm{S}_k\bm{c}_{k,n}^\prime)$
    \EndFor
\EndWhile
\State Compute $\{\bm{w}_k = \bm{S}_k \bm{c}_k\}$
\State Return $\{\bm{w}_k\}$
\end{algorithmic}
\label{alg:meta}
\end{algorithm}
%

%%%%%%%%%%%%%%%%%%%%%%%%%%%%%%%%%%%%%%%%%%%%%%%%%%%%%%%%%%%%%%%%%%%%%%%%%%%%%%%%%%%%%%%%%
\section{Experiments}
\label{sec:exp}
In this section, we evaluate our model for few-shot image captioning on three widely used datasets under different settings with multiple metrics. We first describe the setup of our experiments in Section~\ref{subsec:setup} and then present and analyze the experimental results in Section~\ref{subsec:result}.

% ------------------------------------------------------------------------------------
\subsection{Experimental Setup}
\label{subsec:setup}
\noindent \textbf{Datasets.} To evaluate the multi-modal few-shot meta-learning framework, the datasets need to be organized into sequences of tasks following the episode scheme~\cite{santoro2016meta}. For meta-train, we use the training set of MSCOCO-2017 captioning dataset~\cite{lin2014microsoft} in which each image has around 5 crowd-sourced captions, and we construct tasks in an $N$-way $K$-shot fashion based on the $N$ object categories present in the images. For meta-test, we use 1). the validation set of MSCOCO-2017 captioning dataset, 2). Flickr8k~\cite{hodosh2013framing} consisting of 8,000 images, and 3). Flickr30k~\cite{young2014image} consisting of 31,000 images, in all of which each image is also annotated with 5 crowd-sourced captions. For Flickr8k and Flickr30k datasets, we use pretrained RAM~\cite{zhang2023recognize} to tag each image and classify the image into one of the 80 categories provided by MSCOCO-2017 dataset using exact match and/or fuzzy match~\cite{chan2023ic}. Note that the object categories are split into disjoint groups between the meta-train set and meta-test set. 
% The preprocessing procedure and category split are provided in the \textit{supplementary material}.

\noindent \textbf{Metrics.} We adopt traditional n-gram-based scores including BLEU~\cite{papineni2002bleu}, METEOR~\cite{agarwal2008meteor}, ROUGE~\cite{rouge2004package} and CIDEr~\cite{vedantam2015cider}. In addition, we also adopt two other automated measures of caption quality, 
\ie, \textit{CLIP Recall} and \textit{Reference Content Coverage}~\cite{chan2023ic}, which overcome the limitations of the n-gram-based scores. Specifically, we utilize CLIP~\cite{radford2021learning} as a recall model to estimate the specificity of our generated captions. This involves computing the CLIP embeddings for each image and its corresponding caption, calculating the CLIP-Score between each image and all other generated captions, and subsequently determining the mean reciprocal rank (MRR) and recall at 1, 5, and 10. Higher MRR values indicate more discriminative and detailed captions within the meta-test set. 
% Beyond individual caption specificity, assessed via the CLIP recall measure, 
We also quantify the extent of total information from the image kept in the generated caption by computing the percentage of exact and fuzzy matches of nouns and verbs between the generated captions and the ground-truth captions. 
% More information can be found in the \textit{supplementary material} and~\cite{chan2023ic}.

\noindent \textbf{Train and inference.} We evaluate two distinct training procedures: (i) traditional mini-batched training, referred to as \textit{non-episodic}~\cite{mokady2021clipcap}; (ii) our proposed model that involves training on batches of multi-modal tasks, designated as \textit{episodic}~\cite{santoro2016meta,finn2017model,najdenkoska2022meta}. These procedures are assessed within two contexts determined by the domain shift during training and inference: (i) cross-domain multi-modal few-shot learning, where the training and test datasets stem from disparate distributions, \ie, trained on MSCOCO-2017 (both episodically and non-episodically), and evaluated on Flickr8k and Flickr30k; (ii) in-domain multi-modal few-shot learning, where the training and test partitions both originate from MSCOCO-2017. 
% This suggests that the training and test partitions, namely meta-train and meta-test tasks, are constructed from the identical multimodal few-shot dataset.

\noindent \textbf{Implementation details.}
The vision encoder is constructed utilizing CLIP~\cite{radford2021learning}, with the Vision Transformer (ViT/B-32)~\cite{dosovitskiy2020image} acting as the foundational model, generating visual features of 512 dimensions. The language model is GPT-2~\cite{radford2019language}, with word embeddings of 768 dimensions. Both of the vision and language models are frozen during training. The learnable visual prefix is empirically determined to be of length $\{1, 1, 4\}$ with dimensions of 768 for three steps of CoT. The meta-adaptor is initialized using Xavier uniform initialization~\cite{glorot2010understanding}. For the meta-learning hyperparameters, we empirically establish one inner-loop gradient-updates with a learning rate of 0.01. The meta-update is executed with AdamW~\cite{loshchilov2018decoupled} with a meta-learning rate of 0.001 and 32 tasks in each meta-batch. 
% All hyperparameters are optimized using the query sets in the meta-train partition, which serves as a validation set.

%%%%%%%%%%%%%%%%%%%%%%%%%%%%%%%%%%%%%%%%%%%%%%%%%%%%%%%%%%%%%%%%%%%%%%%%%%%%%%%%%%%%%%%%%
\subsection{Results \& Discussion}
\label{subsec:result}

\begin{table*}[!htb]
% \scriptsize
\centering
\setlength\tabcolsep{1.8pt}
\caption{Quantitative evaluation on MSCOCO, Flickr8k and Flickr30k using BLEU metrics.}
\begin{tabular}{llcccccccccccccccc}
\toprule
        & & \multicolumn{4}{c}{\textbf{BLEU@1}} & \multicolumn{4}{c}{\textbf{BLEU@2}} & \multicolumn{4}{c}{\textbf{BLEU@3}} & \multicolumn{4}{c}{\textbf{BLEU@4}}  \\
        \cmidrule(lr){3-6} \cmidrule(lr){7-10} \cmidrule(lr){11-14} \cmidrule(lr){15-18} 
        & & \multicolumn{2}{c}{\textbf{2-way}} & \multicolumn{2}{c}{\textbf{5-way}} & \multicolumn{2}{c}{\textbf{2-way}} & \multicolumn{2}{c}{\textbf{5-way}} & \multicolumn{2}{c}{\textbf{2-way}} & \multicolumn{2}{c}{\textbf{5-way}} & \multicolumn{2}{c}{\textbf{2-way}} & \multicolumn{2}{c}{\textbf{5-way}} \\
        \cmidrule(lr){3-4} \cmidrule(lr){5-6} \cmidrule(lr){7-8} \cmidrule(lr){9-10} \cmidrule(lr){11-12} \cmidrule(lr){13-14} \cmidrule(lr){15-16} \cmidrule(lr){17-18}  
        \textbf{Datasets}          & \textbf{Methods} & 1-shot & 5-shots & 1-shot & 5-shot & 1-shots & 5-shot &  1-shots  & 5-shot  & 1-shots & 5-shot  & 1-shots & 5-shot  & 1-shots & 5-shot  & 1-shots & 5-shots \\
        \midrule
        \multirow{3}{*}{MSCOCO}    & ClipCap       & 22.73       & 26.49       & 22.59       & 32.27       & 4.84           & 7.53        & 4.81        & 13.26       & 0.28       & 2.07       & 0.29       & 3.78       & 0.00       & 0.43       & 0.06       & 1.13 \\
                                   & Meta-Mapper   & 29.92       & 35.12       & 35.04       & \bfy{34.65} & 11.70          & 16.37       & 16.54       & 15.41       & 2.89       & 5.03       & \bfy{5.98} & 4.78       & 1.31       & 1.88       & \bfy{1.91} & \bfy{1.14} \\
                                   & \textbf{Ours} & \bfy{32.38} & \bfy{37.21} & \bfy{35.59} & 34.56       & \bfy{14.16}    & \bfy{17.81} & \bfy{16.59} & \bfy{15.68} & \bfy{4.29} & \bfy{5.91} & 5.44       & \bfy{4.94} & \bfy{1.68} & \bfy{2.26} & 1.74       & 0.86 \\
        \midrule
        \multirow{3}{*}{Flickr8k}  & ClipCap       & 22.54       & 23.17       & 22.66       & 27.09       & 5.36       & 4.69        & 5.55        & 8.81        & 0.70       & 0.66       & 0.74       & 2.53       & 0.46       & 0.00       & 0.18       & 0.49 \\
                                   & Meta-Mapper   & 27.59       & 28.39       & 29.77       & \bfy{29.35} & 9.01       & 9.76        & 11.57       & 10.65       & 2.45       & 2.42       & 3.08       & \bfy{2.79} & 0.72       & \bfy{0.73} & 0.87       & \bfy{0.79} \\
                                   & \textbf{Ours} & \bfy{29.68} & \bfy{29.56} & \bfy{30.69} & \bfy{29.35} & \bfy{9.79} & \bfy{10.73} & \bfy{12.60} & \bfy{10.77} & \bfy{2.96} & \bfy{2.85} & \bfy{3.68} & 2.53       & \bfy{0.87} & 0.56       & \bfy{1.22} & 0.67 \\
        \midrule
        \multirow{3}{*}{Flickr30k} & ClipCap       & 27.07       & 26.79       & 25.18       & 29.24       & 10.75       & 7.79        & 6.68        & 9.99        & \bfy{2.75} & 1.37       & 0.88       & 2.27       & \bfy{1.61} & 0.42       & 0.36       & 0.48 \\
                                   & Meta-Mapper   & 31.39       & 31.00       & 31.67       & \bfy{31.79} & \bfy{11.84} & 11.23       & 12.38       & \bfy{12.57} & 2.58       & 2.45       & 2.64       & 2.48       & 1.32       & 0.75       & 0.85       & 0.70 \\
                                   & \textbf{Ours} & \bfy{32.25} & \bfy{31.89} & \bfy{32.05} & 31.61       & 11.83       & \bfy{13.22} & \bfy{13.44} & 12.47       & 2.41       & \bfy{4.80} & \bfy{3.61} & \bfy{3.26} & 1.30       & \bfy{2.20} & \bfy{1.49} & \bfy{0.84} \\
\bottomrule
\end{tabular}
\label{tab:bleu}
\end{table*}
\begin{table*}[!htb]
% \scriptsize
\centering
\setlength\tabcolsep{2.5pt}
\caption{Quantitative evaluation on MSCOCO, Flickr8k and Flickr30k using ROUGE, CIDEr and MAUVE.}
\begin{tabular}{llcccccccccccc}
\toprule
        & & \multicolumn{4}{c}{\textbf{ROUGE}} & \multicolumn{4}{c}{\textbf{CIDEr}} & \multicolumn{4}{c}{\textbf{MAUVE}}  \\
        \cmidrule(lr){3-6} \cmidrule(lr){7-10} \cmidrule(lr){11-14}
        & & \multicolumn{2}{c}{\textbf{2-way}} & \multicolumn{2}{c}{\textbf{5-way}} & \multicolumn{2}{c}{\textbf{2-way}} & \multicolumn{2}{c}{\textbf{5-way}} & \multicolumn{2}{c}{\textbf{2-way}} & \multicolumn{2}{c}{\textbf{5-way}} \\
        \cmidrule(lr){3-4} \cmidrule(lr){5-6} \cmidrule(lr){7-8} \cmidrule(lr){9-10} \cmidrule(lr){11-12} \cmidrule(lr){13-14} 
        \textbf{Datasets}          & \textbf{Methods} & 1-shot & 5-shots & 1-shot & 5-shot & 1-shots & 5-shot &  1-shots  & 5-shot  & 1-shots & 5-shot  & 1-shots & 5-shot \\
        \midrule
        \multirow{3}{*}{MSCOCO}    & ClipCap       & 22.51       & 26.05       & 22.27       & 31.24       & 8.78        & 22.64       & 9.19        & 55.00       & 1.96        & 35.28       & 2.00        & 27.71 \\
                                   & Meta-Mapper   & 29.05       & 33.83       & 34.07       & 33.27       & 45.53       & 69.06       & \bfy{69.44} & \bfy{65.41} & 58.38       & \bfy{58.39} & \bfy{46.50} & 38.20 \\
                                   & \textbf{Ours} & \bfy{31.00} & \bfy{35.43} & \bfy{34.54} & \bfy{33.37} & \bfy{63.75} & \bfy{80.61} & 65.14       & 65.14       & \bfy{59.01} & 56.90       & 38.24       & \bfy{44.40} \\
        \midrule
        \multirow{3}{*}{Flickr8k}  & ClipCap       & 22.75       & 22.94       & 22.66       & 27.29       & 10.16       & 15.04       & 10.05       & 32.62       & 2.73        & 36.95       & 4.42        & 24.22 \\
                                   & Meta-Mapper   & 27.59       & 27.21       & 29.88       & 29.04       & 30.58       & 37.75       & 35.26       & \bfy{37.94} & 63.88       & 66.33       & \bfy{42.43} & \bfy{46.62} \\
                                   & \textbf{Ours} & \bfy{28.94} & \bfy{28.52} & \bfy{30.63} & \bfy{29.09} & \bfy{35.55} & \bfy{41.24} & \bfy{40.83} & 36.65       & \bfy{74.04} & \bfy{83.60} & 32.99       & 40.08 \\
        \midrule
        \multirow{3}{*}{Flickr30k} & ClipCap       & 27.90       & 26.93       & 25.21       & 29.06       & 14.86       & 19.14       & 10.46       & 29.46       & 2.20        & 17.20       & 5.10        & 22.70 \\
                                   & Meta-Mapper   & 31.68       & 31.31       & 31.64       & \bfy{31.73} & 33.34       & 37.91       & 31.59       & 34.18       & 52.30       & \bfy{75.58} & 43.05       & 36.05 \\
                                   & \textbf{Ours} & \bfy{32.67} & \bfy{32.35} & \bfy{31.77} & 31.50       & \bfy{35.76} & \bfy{51.48} & \bfy{40.11} & \bfy{34.49} & \bfy{69.49} & 68.96       & \bfy{48.75} & \bfy{44.24} \\
\bottomrule
\end{tabular}
\label{tab:rouge}
\end{table*}
\begin{table*}[!htb]
% \scriptsize
\centering
\setlength\tabcolsep{1.8pt}
\caption{Quantitative evaluation on MSCOCO, Flickr8k and Flickr30k using CLIP Recall and Reference Content Coverage.}
\begin{tabular}{llcccccccccccccccc}
\toprule
        & & \multicolumn{4}{c}{\textbf{MRR}} & \multicolumn{4}{c}{\textbf{Recall@1}} & \multicolumn{4}{c}{\textbf{Recall@5}} & \multicolumn{4}{c}{\textbf{Recall@10}}  \\
        \cmidrule(lr){3-6} \cmidrule(lr){7-10} \cmidrule(lr){11-14} \cmidrule(lr){15-18} 
        & & \multicolumn{2}{c}{\textbf{2-way}} & \multicolumn{2}{c}{\textbf{5-way}} & \multicolumn{2}{c}{\textbf{2-way}} & \multicolumn{2}{c}{\textbf{5-way}} & \multicolumn{2}{c}{\textbf{2-way}} & \multicolumn{2}{c}{\textbf{5-way}} & \multicolumn{2}{c}{\textbf{2-way}} & \multicolumn{2}{c}{\textbf{5-way}} \\
        \cmidrule(lr){3-4} \cmidrule(lr){5-6} \cmidrule(lr){7-8} \cmidrule(lr){9-10} \cmidrule(lr){11-12} \cmidrule(lr){13-14} \cmidrule(lr){15-16} \cmidrule(lr){17-18}  
        \textbf{Datasets}          & \textbf{Methods}  & 1-shot & 5-shots & 1-shot & 5-shot & 1-shots & 5-shot & 1-shots & 5-shot & 1-shots & 5-shot & 1-shots & 5-shot & 1-shots & 5-shot & 1-shots & 5-shots \\
        \midrule
        \multirow{3}{*}{MSCOCO}    & ClipCap       & 2.79        & 13.32       & 1.37        & 24.66       & 0.50        & 6.50        & 0.40        & 11.80       & 2.00        & 18.50       & 1.00        & 36.00       & 3.5         & 27.50       & 1.20        & 54.40 \\
                                   & Meta-Mapper   & 31.04       & 45.65       & 30.30       & \bfy{34.26} & 17.00       & 30.00       & 16.80       & \bfy{19.00} & 47.00       & 66.50       & 43.60       & 50.20       & 65.50       & 80.50       & 63.80       & \bfy{70.40} \\
                                   & \textbf{Ours} & \bfy{42.25} & \bfy{56.33} & \bfy{35.03} & 33.69       & \bfy{28.00} & \bfy{42.00} & \bfy{20.60} & 18.20       & \bfy{60.50} & \bfy{76.50} & \bfy{49.40} & \bfy{51.40} & \bfy{75.00} & \bfy{89.50} & \bfy{68.00} & 69.40 \\
        \midrule
        \multirow{3}{*}{Flickr8k}  & ClipCap       & 2.90        & 8.00        & 1.47        & 15.69       & 0.50        & 0.35        & 0.20        & 6.80        & 2.50       & 9.00        & 0.80        & 21.20       & 4.50        & 15.50       & 2.60        & 35.00 \\
                                   & Meta-Mapper   & 17.93       & 26.35       & 18.24       & 21.03       & 7.5         & 13.00       & 8.20        & 10.00       & 29.00      & 39.50       & 25.80       & 29.80       & 41.00       & 60.00       & 39.40       & 46.20 \\
                                   & \textbf{Ours} & \bfy{24.00} & \bfy{39.63} & \bfy{21.84} & \bfy{22.67} & \bfy{10.50} & \bfy{24.50} & \bfy{10.60} & \bfy{11.80} & \bfy{37.5} & \bfy{56.00} & \bfy{32.20} & \bfy{34.20} & \bfy{56.50} & \bfy{71.00} & \bfy{47.60} & \bfy{48.00} \\
        \midrule
        \multirow{3}{*}{Flickr30k} & ClipCap       & 3.35        & 6.82        & 1.59        & 16.75       & 0.50        & 1.50        & 0.20        & 8.00        & 3.50        & 8.00        & 1.60        & 23.6        & 6.50        & 14.00       & 1.80        & 35.00 \\
                                   & Meta-Mapper   & 17.71       & 26.95       & 18.42       & \bfy{23.74} & 7.50        & 14.50       & 9.40        & \bfy{11.20} & 25.00       & 41.00       & 26.40       & \bfy{36.80} & 36.50       & 53.50       & 39.80       & \bfy{52.40} \\
                                   & \textbf{Ours} & \bfy{28.40} & \bfy{35.14} & \bfy{23.73} & 23.26       & \bfy{16.00} & \bfy{18.00} & \bfy{13.00} & 11.00       & \bfy{39.00} & \bfy{58.50} & \bfy{33.20} & 34.60       & \bfy{57.00} & \bfy{73.00} & \bfy{47.20} & 51.00 \\
\bottomrule
\end{tabular}
\label{tab:clipcover}
\end{table*}
\begin{table*}[!htb]
% \scriptsize
\centering
\setlength\tabcolsep{1.8pt}
\caption{Quantitative evaluation on MSCOCO, Flickr8k and Flickr30k using exact and fuzzy matches of noun and verb.}
\begin{tabular}{llcccccccccccccccc}
\toprule
        & & \multicolumn{4}{c}{\textbf{ECACT-NOUN}} & \multicolumn{4}{c}{\textbf{ECACT-VERB}} & \multicolumn{4}{c}{\textbf{FUZZY-NOUN}} & \multicolumn{4}{c}{\textbf{FUZZY-VERB}}  \\
        \cmidrule(lr){3-6} \cmidrule(lr){7-10} \cmidrule(lr){11-14} \cmidrule(lr){15-18} 
        & & \multicolumn{2}{c}{\textbf{2-way}} & \multicolumn{2}{c}{\textbf{5-way}} & \multicolumn{2}{c}{\textbf{2-way}} & \multicolumn{2}{c}{\textbf{5-way}} & \multicolumn{2}{c}{\textbf{2-way}} & \multicolumn{2}{c}{\textbf{5-way}} & \multicolumn{2}{c}{\textbf{2-way}} & \multicolumn{2}{c}{\textbf{5-way}} \\
        \cmidrule(lr){3-4} \cmidrule(lr){5-6} \cmidrule(lr){7-8} \cmidrule(lr){9-10} \cmidrule(lr){11-12} \cmidrule(lr){13-14} \cmidrule(lr){15-16} \cmidrule(lr){17-18}  
        \textbf{Datasets}          & \textbf{Methods} & 1-shot & 5-shots & 1-shot & 5-shot & 1-shots & 5-shot &  1-shots  & 5-shot  & 1-shots & 5-shot  & 1-shots & 5-shot  & 1-shots & 5-shot  & 1-shots & 5-shots \\
        \midrule
        \multirow{3}{*}{MSCOCO}    & ClipCap       & 6.42        & 8.35        & 6.01        & 23.64       & 0.25       & 5.50        & 2.73        & 11.30       & 27.82       & 38.54       & 29.03       & 54.41       & 12.83       & 24.94       & 31.39       & 30.64 \\
                                   & Meta-Mapper   & 21.22       & 26.30       & 28.94       & \bfy{28.36} & 5.75       & 9.17        & 12.93       & \bfy{12.17} & 52.76       & 55.92       & 58.70       & \bfy{57.30} & 28.36       & 30.40       & 32.87       & \bfy{34.39} \\
                                   & \textbf{Ours} & \bfy{27.97} & \bfy{30.65} & \bfy{30.48} & 28.11       & \bfy{9.25} & \bfy{11.17} & \bfy{13.33} & 12.10       & \bfy{56.22} & \bfy{59.03} & \bfy{58.94} & 57.26       & \bfy{30.88} & \bfy{32.45} & \bfy{34.32} & 31.24 \\
        \midrule
        \multirow{3}{*}{Flickr8k}  & ClipCap       & 7.92        & 9.50        & 9.32        & 20.45       & 0.00       & 4.25        & 2.30        & 5.80       & 38.07       & 42.83       & 38.59       & 50.18       & 16.26       & 32.68       & 33.91       & 34.86 \\
                                   & Meta-Mapper   & 18.22       & 23.97       & 23.95       & \bfy{23.53} & 7.25       & 9.00        & 7.90        & 9.00       & 48.44       & 53.26       & 52.70       & \bfy{52.82} & 37.95       & 40.78       & 38.87       & \bfy{39.35} \\
                                   & \textbf{Ours} & \bfy{19.75} & \bfy{25.73} & \bfy{25.76} & 23.48       & \bfy{7.75} & \bfy{11.25} & \bfy{11.10} & \bfy{9.90} & \bfy{50.46} & \bfy{54.40} & \bfy{54.38} & 52.39       & \bfy{39.24} & \bfy{43.16} & \bfy{41.74} & 38.14 \\
        \midrule
        \multirow{3}{*}{Flickr30k} & ClipCap       & 18.74       & 13.18       & 13.71       & 19.89       & 0.17        & 6.75        & 2.80        & 9.60        & 44.80       & 43.44       & 41.00       & 50.74       & 14.44       & 37.12       & 35.42       & 40.06 \\
                                   & Meta-Mapper   & 21.82       & 24.64       & 24.02       & 24.21       & \bfy{12.00} & 11.75       & 11.20       & 10.10       & 51.21       & 52.60       & 53.28       & 53.18       & \bfy{40.56} & 43.96       & 41.79       & \bfy{41.83} \\
                                   & \textbf{Ours} & \bfy{24.82} & \bfy{26.89} & \bfy{26.08} & \bfy{24.33} & 9.50        & \bfy{12.75} & \bfy{12.97} & \bfy{10.20} & \bfy{52.97} & \bfy{55.79} & \bfy{55.26} & \bfy{53.93} & 39.19       & \bfy{44.56} & \bfy{44.42} & 40.89 \\
\bottomrule
\end{tabular}
\label{tab:nounverb}
\end{table*}
\begin{figure*}[!htb]
    \centering
    \includegraphics[width=0.99\linewidth]{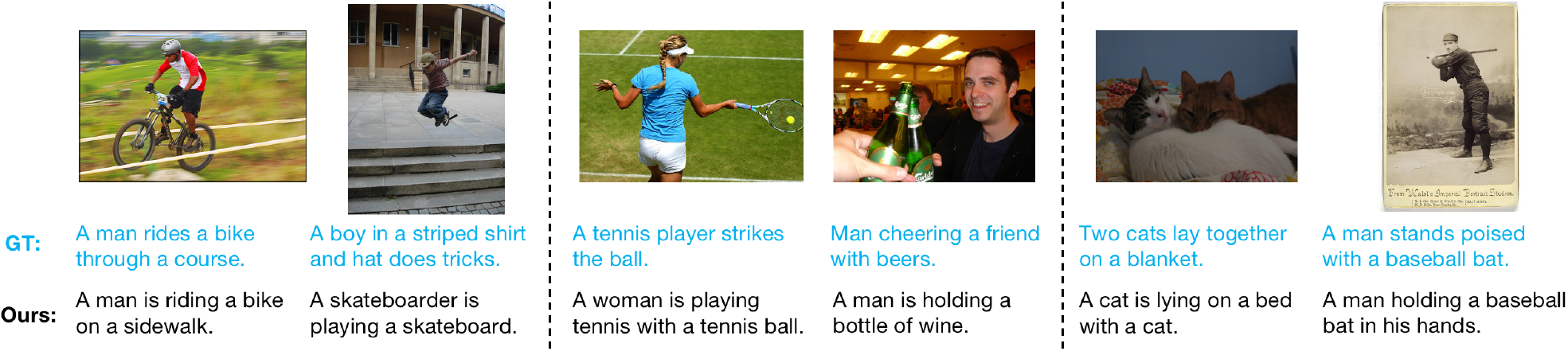}
    \caption{Qualitative evaluation on MSCOCO (left), Flickr8k (middle) and Flickr (right). Our method generate captions that more accurately capture the subject (a \underline{man} in the rightmost image) and the corresponding actions (\underline{\textit{holding} a baseball bat}).}
    \label{fig:qualitative}
\end{figure*}

The experiments undertaken on MSCOCO-2017, Flickr8k and Flickr30k evaluate the performance of the proposed multi-modal chain-of-thought subspace meta-learning in few-shot image captioning. As depicted in Table~\ref{tab:bleu} to Table~\ref{tab:nounverb}, the 2-way and 5-way accuracy in 1-shot and 5-shot across all datasets are compared against two baseline approaches: ClipCap~\cite{mokady2021clipcap} and Meta-Mapper~\cite{najdenkoska2022meta}. Our multi-modal meta-learning method surpass the compared approaches in most cases using different metrics. Compared with Meta-Mapper which also adopts a meta-learning framework but only utilizes a single visual prompt, \ie, global visual prompt, our method demonstrate the superiority of decomposing image captioning process into multiple steps in a CoT manner. Especially, in Table~\ref{tab:nounverb} we notice that even if we do not explicitly extract visual prompt for \textit{verb}, our method still achieves the best performance on the exact and fuzzy vert matches, which indicates that \textit{our visual CoT can utilize the language prior learned in pretrained LLM}. The qualitative results are shown in Figure~\ref{fig:qualitative}.

% This underscores the merit of possessing a meta-learned visual prefix, as opposed to merely transforming the vision encoder output into a prefix for language models. Specifically, the meta-learned prefix has the capacity to amass shared meta-knowledge from related instances within the tasks, which proves crucial in delimiting the search space in a learnable manner, rather than resorting to an engineered task instruction. The inclusion of this adaptable component facilitates the model's adjustment to novel visual concepts, without the necessity for explicit instruction. Moreover, akin to \red{XXX}, a rising trend is noted when the number of shots is augmented, for both the 2-way and 5-way settings, which substantiates that the inclusion of more distinct examples of the task enhances performance.

\noindent \textbf{Non-episodic and episodic scheme.} 
ClipCap also adopts an attention module to align the embedding space of a pretrained LVM encoder and LLM which share the same architectures as ours, but trained in a non-episodic and non-meta-learning manner. However, as shown in  Table~\ref{tab:bleu} to Table~\ref{tab:nounverb}, it performs worst in most cases, which is predominantly ascribed to the discrepancy of schemes during the training and inference stages without a meta-learning framework.
% A significant inference drawn from the evidence provided in Tables~\ref{} and Table~\ref{} is the augmentation of total efficacy via episodic training. This improvement is predominantly ascribed to the synchronization of circumstances during the training and inference stages. To provide further detail, the multimodal few-shot datasets employed during the inference stage are structured into distinct tasks. Therefore, implementing a corresponding strategy during the training stage, wherein the model is subjected to these tasks, culminates in a remarkable improvement in performance.

%
\begin{table}[!htb]
\scriptsize
\centering
\setlength\tabcolsep{2.5pt}
\caption{Ablative study of each component in our model.}
\begin{tabular}{ccccccc}
\toprule
\textbf{Subspace}             & \textbf{Sub}                  & \textbf{Obj}                  & \textbf{BLUE@1}               & \textbf{BLUE@2}               & \textbf{BLUE@3}               & \textbf{BLUE@4}               \\
\midrule
        \xmark           &        \xmark            &          \xmark         &           27.59            &         9.01           &          2.45            &         0.72             \\
        \cmark           &        \xmark            &          \xmark         &           28.60            &         9.44           &          2.68            &         0.79             \\
        \cmark           &        \cmark            &          \xmark         &           29.14            &         9.57           &          2.74            &         0.83             \\
        \cmark           &        \xmark            &          \cmark         &           29.10            &         9.52           &          2.70            &         0.80             \\
       \cmark            &        \cmark            &          \cmark         &          29.68             &        9.79            &          2.96            &         0.87              \\
% \multicolumn{1}{l}{} & \multicolumn{1}{l}{} & \multicolumn{1}{l}{} & \multicolumn{1}{l}{} & \multicolumn{1}{l}{} & \multicolumn{1}{l}{} & \multicolumn{1}{l}{} \\ 
\bottomrule
\end{tabular}
\label{tab:abl}
\end{table}

\noindent \textbf{In-domain and cross-domain learning.} As our model is trained on MSCOCO-2017 training set, but evaluated on in-domain set, \ie, the validation set of MSCOCO-2017, and cross-domain set, \ie, Flickr8k and Fickr30k. We observe that 1). the performance on MSCOCO is better than that on Flickr8k and Flickr30k, which is consist with our expectation; and 2). our method still achieves superior performance than ClipCap and Meta-Mapper under cross-domain settings, which indictates the robustness of our proposed CoT captioning procedure.
% A detailed nnalysis of Tables~\ref{} and~\ref{} illuminate the complexities inherent in the more rigorous cross-domain multimodal few-shot scenario. At its core, meta-train with the MSCOCO-2017 captioning dataset enhances the model's capacity to correlate visual concepts with more elaborate semantic descriptions, in contrast to training on multimodal few-shot datasets, which are characterized by more rudimentary captions. As a result, a model meta-trained on MSCOCO-2017 exhibits reduced accuracy from a quantitative perspective when transferred to a Flickr-based dataset. Yet, in a significant number of cases, the sentences generated are of higher qualitative value and exhibit greater detail, as we discuss in the following paragraph.

\noindent
% Figure~\ref{} displays a several query images, each paired with generated captions during the inference phase, as produced by the most successful version of our approach according to Tables~\ref{} and Table~\ref{}. \blue{The aptitude of the multimodal meta-learner in correlating visual concepts with verbal representations is clearly demonstrated: the model effectively associates the visual components in the image not only with the phrase 'electric guitar', as specified in the ground-truth, but also with the term 'player'. This suggests that the model is efficiently directed by the meta-learned visual prefix to harness supplementary information from the image, which may not be explicitly contained in the ground-truth.} An important observation from these instances is the existence of variances between the generated answers and the ground-truth, which are subject to penalties during evaluation since we only recognize exact word matches as correct to ensure comparability with Frozen. Nonetheless, the evaluation should ideally determine whether the generated sentence broadly corresponds with the image, without the necessity of ground-truth references~\cite{hessel2021clipscore}, an aspect left for future investigation.

%%%%%%%%%%%%%%%%%%%%%%%%%%%%%%%%%%%%%%%%%%%%%%%%%%%%%%%%%%%%%%%%%%%%%%%%%%%%%%%%%%%%%%%%%
\section{Ablation Study}
\label{sec:abl}
In this section, we evaluate the effectiveness of each component in our proposed model and the ablative results are shown in Table~\ref{tab:abl}. All experiments are trained on MSCOCO-2017 and assessed on Flickr8k under 2-way 1-shot settings. From Table~\ref{tab:abl} we note that adding subspace, subject prompot and object prompt is beneficial for the final generated captions. And we also notice that combining subject prompt and object prompt together to form an intact SVO chain further boosts the performance.

%%%%%%%%%%%%%%%%%%%%%%%%%%%%%%%%%%%%%%%%%%%%%%%%%%%%%%%%%%%%%%%%%%%%%%%%%%%%%%%%%%%%%%%%%
\section{Conclusion}
\label{sec:con}
In this paper, we introduce a chain-of-thought meta-learning to closely emulate human image description processes. Moreover, we propose to learn the model's meta-parameters uniquely for each CoT phase within separate subspaces to minimize interference. Our methodology is assessed on prevalent image captioning datasets within few-shot contexts and outperforms benchmark methods across different datasets when evaluated using diverse metrics.

\bibliographystyle{IEEEtran}
\bibliography{ref}

% \bibliographystyle{IEEEtran}

% \bibitem{ref1}
% {\it{Mathematics Into Type}}. American Mathematical Society. [Online]. Available: https://www.ams.org/arc/styleguide/mit-2.pdf

% \bibitem{ref2}
% T. W. Chaundy, P. R. Barrett and C. Batey, {\it{The Printing of Mathematics}}. London, U.K., Oxford Univ. Press, 1954.

% \bibitem{ref3}
% F. Mittelbach and M. Goossens, {\it{The \LaTeX Companion}}, 2nd ed. Boston, MA, USA: Pearson, 2004.

% \bibitem{ref4}
% G. Gr\"atzer, {\it{More Math Into LaTeX}}, New York, NY, USA: Springer, 2007.

% \bibitem{ref5}M. Letourneau and J. W. Sharp, {\it{AMS-StyleGuide-online.pdf,}} American Mathematical Society, Providence, RI, USA, [Online]. Available: http://www.ams.org/arc/styleguide/index.html

% \bibitem{ref6}
% H. Sira-Ramirez, ``On the sliding mode control of nonlinear systems,'' \textit{Syst. Control Lett.}, vol. 19, pp. 303--312, 1992.

% \bibitem{ref7}
% A. Levant, ``Exact differentiation of signals with unbounded higher derivatives,''  in \textit{Proc. 45th IEEE Conf. Decis.
% Control}, San Diego, CA, USA, 2006, pp. 5585--5590. DOI: 10.1109/CDC.2006.377165.

% \bibitem{ref8}
% M. Fliess, C. Join, and H. Sira-Ramirez, ``Non-linear estimation is easy,'' \textit{Int. J. Model., Ident. Control}, vol. 4, no. 1, pp. 12--27, 2008.

% \bibitem{ref9}
% R. Ortega, A. Astolfi, G. Bastin, and H. Rodriguez, ``Stabilization of food-chain systems using a port-controlled Hamiltonian description,'' in \textit{Proc. Amer. Control Conf.}, Chicago, IL, USA,
% 2000, pp. 2245--2249.

% \end{thebibliography}

\newpage

 

% \end{IEEEbiography}



\vfill

\end{document}